\documentclass[runningheads]{llncs}

\usepackage[T1]{fontenc}
\usepackage{graphicx}
\usepackage{hyperref}
\usepackage{booktabs}
\usepackage{multirow}
\usepackage{xcolor}
\usepackage{caption} 
\usepackage[symbol]{footmisc}
\usepackage{stfloats}
\usepackage[caption=false,font=normalsize]{subfig}
\usepackage{cite}
\usepackage{setspace}

\title{Unmasking Performance Gaps: A Comparative Study of Human Anonymization and Its Effects on Video Anomaly Detection
}

\author{Sara Abdulaziz \and
Egor Bondarev}

\authorrunning{S. Abdulaziz and E. Bondarev}

\institute{Eindhoven University of Technology, 5612 AP Eindhoven, The Netherlands
\email{\{s.e.a.m.abdulaziz, e.bondarev\}@tue.nl}\\
}

\begin{document}

\maketitle

\begin{abstract}
Advancements in deep learning have improved anomaly detection in surveillance videos, yet they raise urgent privacy concerns due to the collection of sensitive human data. In this paper, we present a comprehensive analysis of anomaly detection performance under four human anonymization techniques, including blurring, masking, encryption, and avatar replacement, applied to the UCF-Crime dataset \cite{sultani2018real}. We evaluate four anomaly detection methods, MGFN \cite{chen2023mgfn}, UR-DMU \cite{zhou2023dual_urdmu}, BN-WVAD \cite{zhou2024batchnorm}, and PEL4VAD \cite{pu2024learning_pel4vad}, on the anonymized UCF-Crime to reveal how each method responds to different obfuscation techniques. Experimental results demonstrate that anomaly detection remains viable under anonymized data, and is dependent on the algorithmic design and the learning strategy. For instance, under certain anonymization patterns, such as encryption and masking, some models inadvertently achieve higher AUC performance compared to raw data, due to the strong responsiveness of their algorithmic components to these noise patterns. These results highlight the algorithm-specific sensitivities to anonymization and emphasize the trade-off between preserving privacy and maintaining detection utility. Furthermore, we compare these conventional anonymization techniques with the emerging privacy-by-design solutions, highlighting an often overlooked trade-off between robust privacy protection and utility flexibility. Through comprehensive experiments and analyses, this study provides a compelling benchmark and insights into balancing human privacy with the demands of anomaly detection.
\end{abstract}

\section{Introduction}
Privacy-preserving computer vision has emerged as an active research area with the growing emphasis on privacy in surveillance systems due to the recent regulations, such as GDPR and AI Act. Various anonymization techniques have been developed to mitigate these privacy risks while retaining the utility of perception tasks \cite{dave2022spact,fioresi2023ted,shifa2020skin,triess2024exploring,yan2020image}. Traditional anonymization techniques, such as human blurring, pixelation, encryption, masking, and avatar replacement \cite{climent2021protection}, offer basic levels of privacy protection but often degrade the performance of the utility tasks. More advanced techniques, such as privacy-by-design solutions \cite{dave2022spact,li2023stprivacy,fioresi2023ted}, have gained traction in recent years, as they promise stronger privacy protection with minimal utility loss.

While privacy-by-design solutions demonstrate success in complex vision tasks, such as human behavior analysis \cite{dave2022spact,li2023stprivacy} and anomaly detection \cite{fioresi2023ted}, they remain constrained by a number of limitations. Most notably, such solutions result in a severe degradation of scene quality, which prevents visual monitoring, a crucial component of real-world surveillance applications. Additionally, these solutions are often inflexible to generic image processing algorithms, limiting their applicability to a narrow range of use cases. Despite the advancements in privacy-by-design solutions, there remains a lack of systematic comparison between these solutions and the conventional anonymization techniques, particularly on complex vision tasks, such as anomaly detection.

The complexity of video analysis methods within surveillance applications varies significantly. Fundamental tasks such as object detection \cite{object_detection}, tracking \cite{object_tracking}, and re-identification \cite{reid} primarily focus on identifying and following objects or individuals, typically relying on spatial and basic temporal information. In contrast, human activity recognition (HAR) \cite{HAR} and video anomaly detection (VAD) \cite{VAD} require a deeper understanding of complex behavioral patterns and subtle visual cues. Therefore, they rely on more sophisticated spatio-temporal feature extraction and analysis, due to their sensitivity to visual variations and contextual information.

Unlike HAR methods that aim to learn behavioral information from extracted features, VAD models aims to distinguish normal from abnormal events using snippet-based features \cite{sultani2018real}. Since VAD is highly sensitive to subtle visual interframe changes, the real implications of anonymization on its performance remain unclear without deeper analysis. In this paper, we state the following research questions: (1) To what extent do recent anomaly detection models rely on sensitive information, such as human body features, in their decision-making? (2) Does the performance drop caused by the conventional anonymization exceed the degradation observed under privacy-by-design solutions? (3) How do different human replacements, such as blurring, encryption, masking, or avatar replacement, influence the anomaly detection performance?

% ==============  
In this paper, we introduce the Anonymized UCF-Crime (AUCF-Crime), a human privacy-preserved version of the original UCF-Crime dataset \cite{sultani2018real}, designed to facilitate research on privacy-preserving anomaly detection. With this dataset, we evaluate four anonymization methods, including blurring, masking, encryption, and avatar replacement, on four anomaly detection methods \cite{chen2023mgfn,zhou2023dual_urdmu,zhou2024batchnorm,pu2024learning_pel4vad}, examining one feature extraction technique \cite{I3D}. To the best of our knowledge, this is the first large-scale study assessing the impact of multiple anonymization techniques on video anomaly detection performance. This study contributes to the ongoing discussion on balancing the privacy protection and visual utility, offering a benchmark for evaluating privacy-preserving strategies in video anomaly detection. The main contributions are summarized as follows:

\begin{itemize}
    \item We present AUCF-Crime, a public anonymized variant of UCF-Crime \cite{sultani2018real} for anomaly detection, incorporating multiple anonymization techniques, such as human-segment blurring, masking, encryption, and avatar replacement.
    \item We quantify how these anonymization techniques affect the anomaly detection performance, demonstrating that modern algorithms primarily rely on learned separability of spatio-temporal features rather than sensitive human attributes.
    \item We compare conventional anonymization techniques with privacy-by-design solutions, highlighting the trade-off between privacy protection and utility flexibility.
\end{itemize}

\section{Related Work}
Privacy-preserving computer vision has emerged as a critical research area to enable vision-based tasks, such as human detection, action recognition, human pose estimation (HPE), and anomaly detection (AD), while protecting human privacy. Existing approaches can be categorized into two main research directions: (1) visual obfuscation and (2) privacy-by-design.

\textbf{Visual Obfuscation} methods aim to conceal human identities in visual data while preserving sufficient scene information for the downstream vision tasks. These methods typically involve blurring, pixelation, blackening, encryption, or replacing individuals with alternative representations, such as 2D avatars \cite{climent2021protection}, 3D avatars\cite{shen2023privacy}, skeletons \cite{zou2022privacy}, or synthetic identities \cite{DeepPrivacy2}.

Several studies have examined the impact of visual obfuscation on utility performance. Cucchiara et al. \cite{cucchiara2024video} explore HAR performance under face and body blurring, demonstrating that knowledge distillation techniques can compensate for privacy-induced data loss, maintaining action recognition accuracy. The findings suggest that fine-grained human details are not always necessary for HAR tasks. Another study by Yan et al. \cite{yan2020image} focuses on HAR performance on segmented and bounding-box-obfuscated images, concluding that contextual cues play a more significant role than privacy-sensitive features in action recognition. While these studies have evaluated the effects of varying blurring levels \cite{cucchiara2024video} and differences in human shape representation \cite{yan2020image}, our work focuses on the impact of different human region replacements, including blurring, masking, encryption, and avatar-based substitution.

Beyond HAR, obfuscation has been applied to privacy-preserving behavior analysis. Mishra et al. \cite{mishra2023privacy} extract body skeletons and semantic masks to study behavioral patterns in dementia risk assessment, evaluating performance with and without background context. The results show only a minor performance degradation when human body segments are removed while background information is maintained, reinforcing previous findings that privacy-sensitive attributes may not be essential for certain vision tasks. Additionally, the authors of \cite{climent2021protection,shen2023privacy} propose human-to-avatar replacements as a privacy-preserving solution for video-based monitoring, demonstrating its feasibility as an alternative to traditional obfuscation techniques.

Recent efforts have also investigated the human pose estimation methods under synthetic-based obfuscation. In \cite{triess2024exploring}, the authors compare human-synthesized images, generated by DeepPrivacy2 \cite{DeepPrivacy2}, with conventional obfuscation techniques, such as blurring and pixelation, to highlight the advantages of synthetic anonymization in retaining the performance of pose-related tasks.

\textbf{Privacy-by-Design} approaches integrate privacy protection within the visual processing pipeline, ensuring that the output is inherently privacy-preserving \cite{xu2021audio,hinojosa2022privhar,PA-HMDB51,fioresi2023ted,li2023stprivacy}. These methods typically generate task-specific anonymized visual data by jointly optimizing the anonymization and utility models \cite{PA-HMDB51}. This approach ensures that the generated data remains useful for predefined utility tasks while obfuscating it against known privacy threats.

While privacy-by-design techniques provide stronger privacy guarantees by enabling human-imperceptible anonymization and reduced privacy leakage in the temporal dimension, they still feature some limitations. To begin with, their task-specific nature limits their flexibility, making such techniques unsuitable for general-purpose image processing. In addition, these approaches continue to exhibit utility performance degradation relative to raw data and are not fairly benchmarked against traditional obfuscation techniques, complicating the assessment of their relative efficiency and associated trade-offs. Despite growing interest in privacy-preserving computer vision, existing evaluations remain limited in scope, often focusing on a narrow set of anonymization techniques and vision tasks. In this work, a comprehensive evaluation of multiple privacy-preserving methods is conducted in the context of video anomaly detection.

\section{Evaluation Methods}
In this section, we outline the evaluation methods applied to analyze the anomaly detection performance under anonymization. It comprises two key subsections: the first details the anonymization process applied to produce the AUCF-Crime; the second describes the anomaly detection methods evaluated on the resulting anonymized versions.

\subsection{UCF-Crime Anonymization}
UCF-Crime \cite{sultani2018real} is a large-scale benchmark dataset for real-world video anomaly detection. It contains surveillance videos categorized into 13 anomalous classes, such as robbery, assault, and vandalism, along with a set of normal videos. We anonymize UCF-Crime by segmenting and obfuscating human regions with conventional techniques, such as blurring, blackening, encryption, and avatar replacement \cite{climent2021protection}. It is important to note that human synthesis methods \cite{DeepPrivacy2} are excluded in this study due to their poor temporal consistency on UCF-Crime videos.

\begin{figure*}[h!]
\centering
\subfloat[Raw]{\includegraphics[width=0.2\linewidth, height=0.25\linewidth]{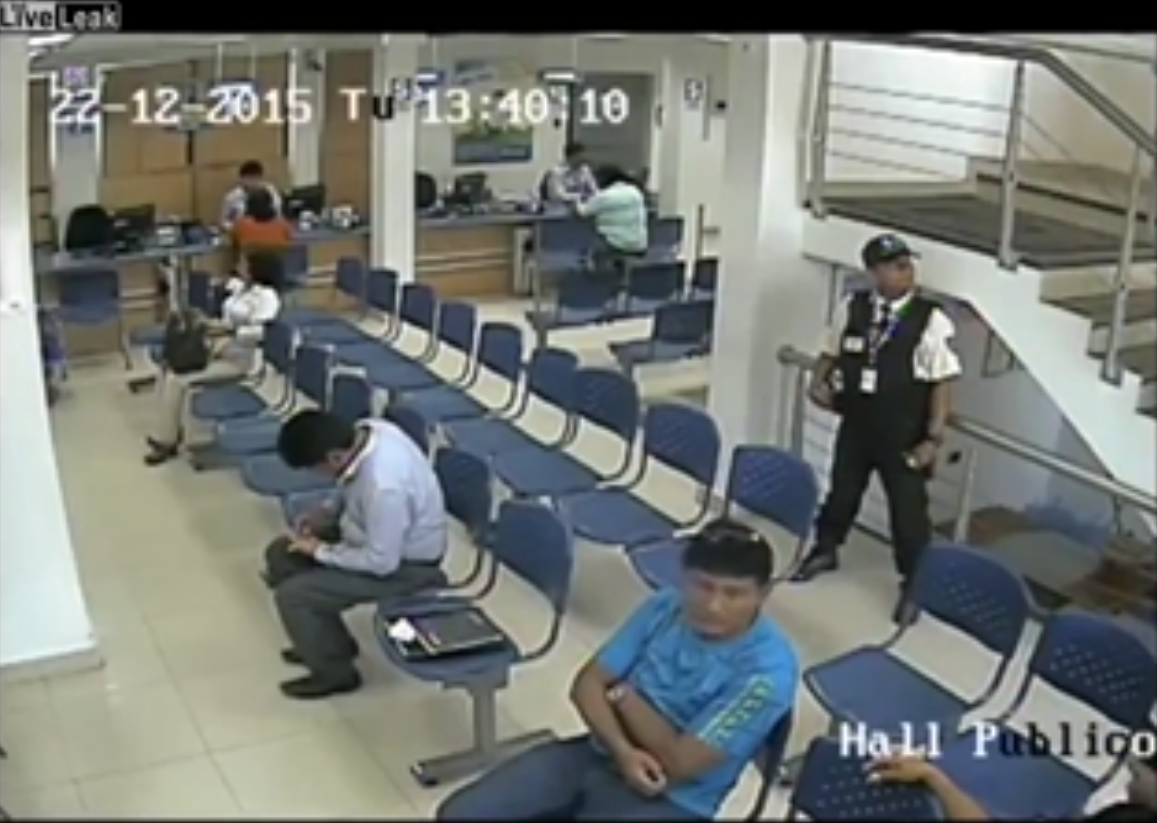}}
\subfloat[IUV map]{\includegraphics[width=0.2\linewidth, height=0.25\linewidth]{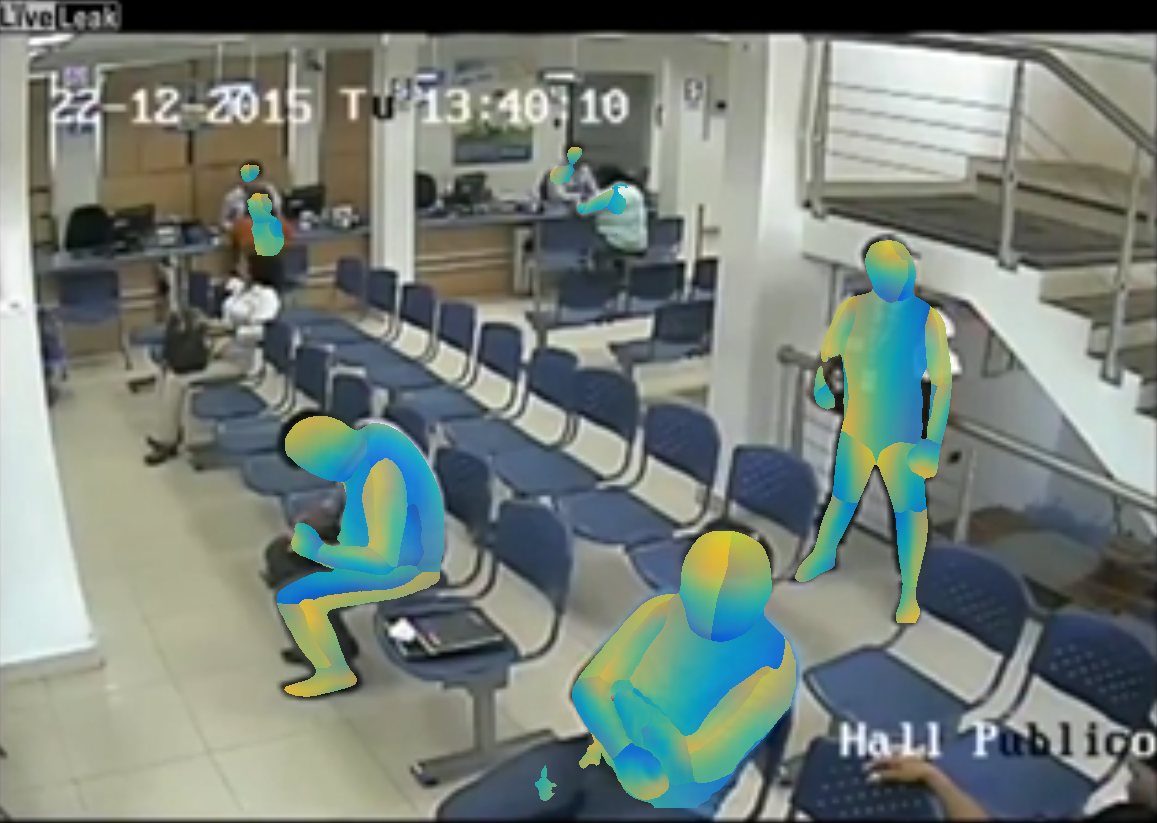}} 
\subfloat[Avatars]{\includegraphics[width=0.2\linewidth, height=0.25\linewidth]{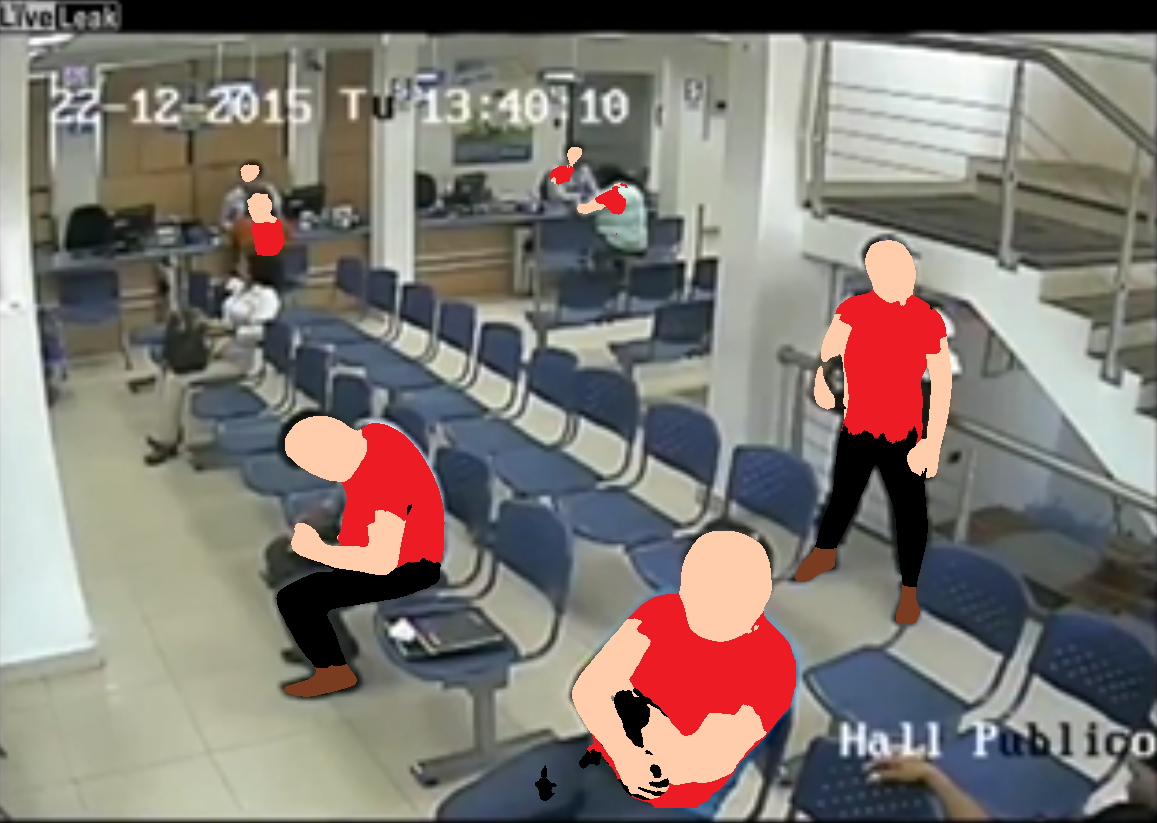}}%\hfill
\subfloat[Atlas]{\includegraphics[width=0.2\linewidth, height=0.25\linewidth]{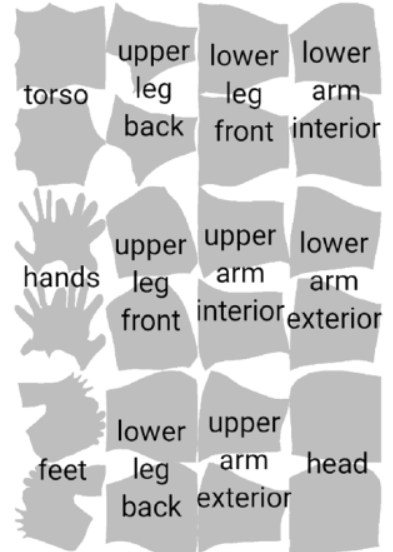}}
\subfloat[Custom texture]{\includegraphics[width=0.2\linewidth, height=0.25\linewidth]{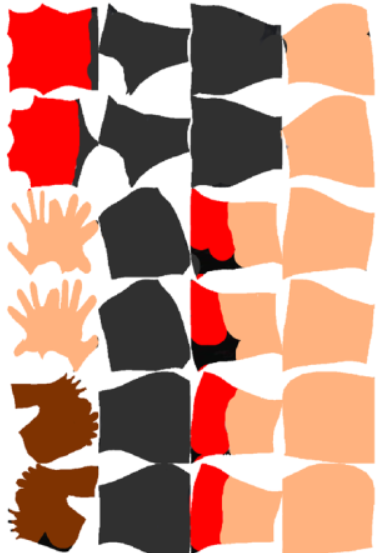}}

\caption{Visualization of the human avatar replacement process. (a) The raw frame (Shooting031). (b) The IUV map inferred by DensePose \cite{guler2018densepose}. (c) The customized texture overlaid on human bodies. This is achieved by mapping the UV coordinates of each body part in (d) to the corresponding XY coordinates of the texture in (e). Figures (d-e) are reprinted from \cite{climent2021protection}. }
\label{fig:avatar-texture}
\end{figure*}

For the blurring, blackening, and encryption, human regions are first segmented by a pre-trained Mask R-CNN model \cite{he2017mask}. The Mask R-CNN provides accurate pixel-level masks for human figures, which serve as input to our anonymization procedures. We apply strong Gaussian blur to each segmented human region, with a kernel size of 101 for blurring obfuscation. The masking method involves completely removing visual details by replacing the pixel values with zero-level pixels. The encryption approach follows a more sophisticated procedure adapted from Shifa et al. \cite{shifa2020skin}. After segmenting human regions, we first encrypt the entire frame independently using AES encryption with Cipher Block Chaining (CBC) mode. Then, the segmented human regions in the original frame are replaced by their corresponding encrypted counterparts from the fully encrypted frame. For avatar replacement, we follow a similar approach to the work of Climent-Pérez et al. \cite{climent2021protection}. First, we apply DensePose \cite{guler2018densepose} to detect human segments and produce IUV maps, which represent the individual body part indices and their location in a UV coordinate space. Then, through the IUV map, a customized avatar texture is applied to replace the original body appearance. The avatar texture discloses the body parts, pose, and actions being performed, while hiding the human identity, as seen in Figure \ref{fig:avatar-texture}.

The resulting AUCF-Crime dataset complies with the original UCF-Crime protocols, and is publicly available to facilitate further research.

\subsection{Anomaly Detection}
Recent works on video anomaly detection heavily rely on I3D features \cite{I3D} in combination with the multiple instance learning (MIL) approach \cite{sultani2018real}, with various modifications to normality modeling, definition of anomaly criterion, and feature-level enhancements \cite{tian2021weakly,chen2023mgfn,zhou2023dual_urdmu,zhou2024batchnorm}. The key similarity between these methods is the primary learning objective, focused on the separability of feature snippets rather than on capturing object-level behavioral and relational details. 

By default, this strategy, which aims at learning the separation between feature representations based on temporal or statistical deviations, tends to circumvent the reliance on sensitive attributes. Therefore, it may allow anomaly detection to be performed under anonymization, without significant performance loss. In this work, the focus is not on the representational power of the I3D features applied in such methods under anonymization, but on the applicability of VAD methods that follow this separation strategy to anonymized videos. To this end, we investigate the effects of human anonymization, explained previously, on the weakly-supervised anomaly detection models MGFN \cite{chen2023mgfn}, UR-DMU \cite{zhou2023dual_urdmu}, BN-WVAD \cite{zhou2024batchnorm}, and PEL4VAD \cite{pu2024learning_pel4vad}, briefly summarized below. 

\textbf{MGFN} \cite{chen2023mgfn} has a dual-branch feature fusion model that combines global and local temporal features for improved anomaly detection. For anomaly snippet selection, the authors rely on feature magnitudes as the primary criterion. To handle the challenge of inconsistent feature magnitudes caused by the scene variations, the authors propose a feature amplification mechanism (FAM) that enhances the discriminativeness of the feature representations. The model is trained based on a proposed magnitude-contrastive loss, by which the network learns to differentiate normal from anomalous segments by reducing intra-class feature magnitude distances and increasing inter-class separation.

\textbf{UR-DMU} \cite{zhou2023dual_urdmu} first enhances the I3D features by capturing long and short range temporal dependencies through a modified self-attention mechanism. The framework has dual memory units which stores prototypical patterns for normal and anomaly. The dual memory units act as external repositories that the network leverages to learn the separation with a dedicated dual memory loss. To handle the noise within the normal data, caused by camera changes or scene variations, the authors propose data uncertainty learning (DUL), which learns a latent space modeled as a Gaussian distribution. The overall modules are jointly optimized under a MIL setting using the video-level labels. 

\textbf{BN-WVAD} \cite{zhou2024batchnorm} incorporates batch normalization statistics to guide weakly supervised anomaly detection. The authors employ BatchNorm layers, which compute a running mean and variance over mini-batches that are dominated by normal snippets. This running mean is used as a statistical representation of normality. The hidden feature of each snippet is compared to the BatchNorm mean, by a Divergence of Feature from Mean (DFM) protocol, filtering out the outliers with high DFM scores. For further separation, a mean-based pull-push loss is proposed. To handle the noise within the normal data, the anomaly classifier is trained on definitive normal snippets from normal videos using snippet-level regression loss. 

\textbf{PEL4VAD} \cite{pu2024learning_pel4vad} focuses on enhancing the separability of anomalous classes through prompt-enhanced learning. The I3D features are first enhanced by a temporal context aggregation (TCA) module, capturing contextual information for enhanced semantic discriminability. Then, semantic prompts representing anomaly nuances are obtained from a knowledge-base, and integrated to boost the discriminative capacity and ensure separability between the anomaly sub-classes. Through the proposed prompt-enhanced learning (PEL), the semantic prompts for anomaly are aligned with the corresponding anomaly features, while non-anomalous features are distanced during training. This injection of prior semantics leads directly to enhanced discriminability.

\section{Experimental Results}
We comply with established evaluation procedures in prior works to ensure equitable comparisons \cite{sultani2018real,dave2022spact,fioresi2023ted}. For VAD performance evaluation, we measure the area under the receiver operating characteristics (AUC), the area under the precision-recall curve (AP), and the false alarm rate (FAR). The scores for the subset with abnormal data only (AUC$_{sub}$ and AP$_{sub}$) and class-wise AUC scores are also applied in the experiments. Throughout this section, the blurring, masking, encryption, and avatar replacement anonymization are noted as HB, HM, HEN, and H2D, respectively.  

\subsection{Implementation Details} 
Following the procedures in prior works \cite{chen2023mgfn,zhou2023dual_urdmu,zhou2024batchnorm,pu2024learning_pel4vad}, we employ the I3D model \cite{I3D} pre-trained on Kinetics-400 to extract the spatio-temporal features for the RGB stream, with a 10-crop augmentation strategy to ensure fair comparison. Features are extracted for 16-frame non-overlapping segments across all anonymized variants (blurring, masking, encryption, and avatar), and across the raw non-anonymized data, to ensure consistency. We follow the training protocols provided by the authors of the SOTA VAD methods MGFN \cite{chen2023mgfn}, UR-DMU \cite{zhou2023dual_urdmu}, BN-WVAD \cite{zhou2024batchnorm}, and PEL4VAD \cite{pu2024learning_pel4vad}, maintaining their recommended hyperparameters, learning rates, and optimization strategies.

\subsection{VAD on Anonymized Videos}
Table \ref{tab:all_metrics} presents the performance comparison of the anomaly detection methods MGFN \cite{chen2023mgfn}, UR-DMU \cite{zhou2023dual_urdmu}, BN-WVAD \cite{zhou2024batchnorm}, and PEL4VAD \cite{pu2024learning_pel4vad}, across selected anonymization techniques. Despite a general decrease in AUC scores when anonymization is applied, the performance of the VAD methods closely correlates with baseline performance on the raw non-anonymized data. This highlights the viability of anomaly detection under anonymization, albeit with robustness varying depending on the chosen anomaly detection algorithm, as detailed below 

\begin{table}[htb!]
    \centering
    \caption{Performance comparison of different anomaly detection methods under anonymization.}
    \renewcommand{\arraystretch}{1}
    \setlength{\tabcolsep}{5pt}  % Adjust column spacing
    \begin{tabular}{c|l|ccccc}
        \hline
        \multirow{2}{*}{\textbf{Method}} & \multirow{2}{*}{\textbf{Protection}} & \multicolumn{5}{c}{\textbf{Metrics (\%)}} \\
        & & AUC & AP & AUC$_{sub}$ & AP$_{sub}$ & FAR ($\downarrow$) \\
        \hline
        \multirow{5}{*}{MGFN \cite{chen2023mgfn}} 
        & \textit{Raw-Reported} & \textit{86.98} & - & -  & -  & -   \\
        \cline{2-7}
        % & Raw-Ours  & 80.05 & 21.03 & 58.11 & 23.22 & 31.29  \\ %experiment repaeted once more for verification 
         & Raw-Ours  & 80.16 & 22.05 & 60.96 & 24.50 & 19.02  \\ 
        & HB  & 80.57 & 20.45 & 59.93 & 23.67 & 26.42  \\
        & HM  & 79.49 & 18.05  & 59.57 & 22.49 & \textbf{2.80} \\
        & HEN  & \textbf{81.20} & \textbf{22.04} & \textbf{60.99} & \textbf{24.86} & 17.65 \\
        & H2D  & 79.46 & 18.70 & 57.66 & 21.56 & 8.69 \\
        \hline
        \multirow{5}{*}{UR-DMU \cite{zhou2023dual_urdmu}} 
        &\textit{Raw-Reported}  & \textit{86.97}  & \textit{35.59} & -  & -  &  \textit{1.05}  \\
        \cline{2-7}
        & Raw-Ours  & \textbf{85.85} & 30.21  & \textbf{69.85} & 32.80 & 2.39  \\
        & HB  & 85.63 & \textbf{35.51} & 69.30 & \textbf{37.80} & 6.11 \\
        & HM  & 85.08 & 35.20 & 66.80 & 36.96 & 2.67 \\
        & HEN  & 81.36  & 22.33 & 62.50 & 25.13 & 2.02 \\
        & H2D  & 84.79 & 29.79 & 68.21 & 29.48 & \textbf{1.77} \\
        \hline
        \multirow{5}{*}{BN-WVAD \cite{zhou2024batchnorm}} 
        & \textit{Raw-Reported}  & \textit{87.24} & \textit{36.26} & \textit{71.71} & \textit{38.13}  &  -  \\
        \cline{2-7}
        & Raw-Ours  & 83.55 & 28.73 & 62.54 & 30.69 & 47.98  \\
        & HB  & 81.72 & 26.29 & 61.26 & 29.02 & 44.74 \\
        & HM  & \textbf{84.52}& \textbf{30.83} & 66.79 & \textbf{32.81} & 99.65 \\
        & HEN  & 81.22 & 25.57  & \textbf{85.13} & 27.55 & 100 \\
        & H2D  & 82.72 &  26.06 & 60.94 & 28.02 & \textbf{35.18} \\
        \hline
        \multirow{5}{*}{PEL4VAD \cite{pu2024learning_pel4vad}} 
        & \textit{Raw-Reported}  & \textit{86.76} & - & - & -  & \textit{0.57}   \\
        \cline{2-7}
        & Raw-Ours & \textbf{85.16} & \textbf{31.07} & 68.23 &  32.86 & 0.35   \\
        & HB  &  84.72 & 30.96  & \textbf{68.53} & \textbf{33.08} & 0.63  \\
        & HM  &  84.53 &  30.66 & 67.62 &  32.44  & \textbf{0.22}  \\
        & HEN  & 82.52   &  23.44  &  61.90  &  25.31  & 0.45   \\
        & H2D  &  84.21  &  27.14   &  66.43  & 29.07   & 0.32   \\
        \hline
    \end{tabular}
    \label{tab:all_metrics}
\end{table}

\begin{figure}
    \centering
    \includegraphics[width=0.5\linewidth]{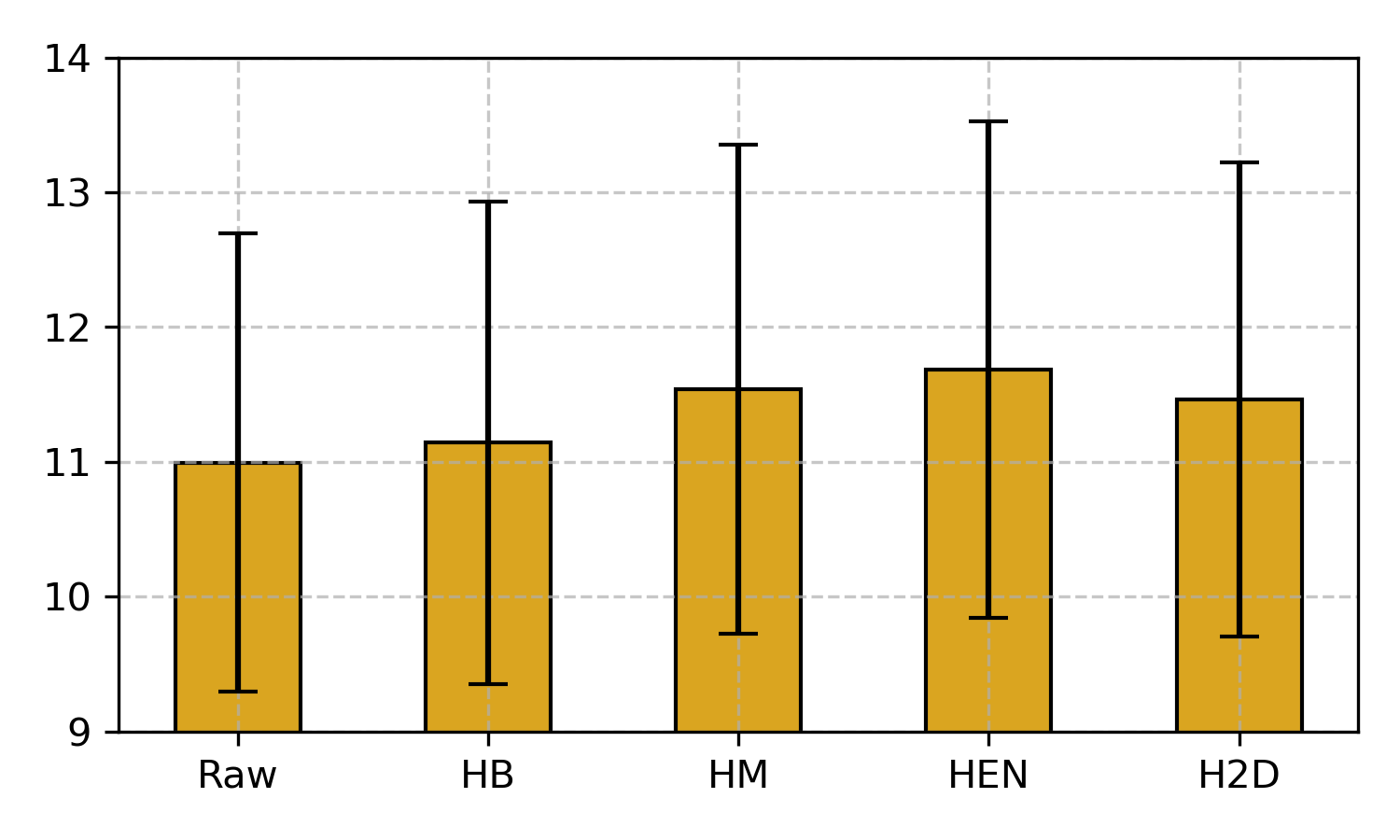}
    \caption{Average magnitudes of raw and anonymized I3D features, computed exclusively for the test snippets corresponding to abnormal annotations. The differences highlight the impact of the anonymization type on feature magnitudes.     
    }
    \label{fig:feature_magnitudes}
\end{figure}

\textbf{MGFN} demonstrates relatively high performance under encryption-based anonymization (HEN), suggesting a sensitivity to the noisy patterns introduced by encryption. These intrinsics are likely attributed to the feature amplification mechanism (FAM), which amplifies feature magnitudes proportionally to a modulated feature norm. Since the feature magnitudes are influenced by the different anonymization techniques, as seen in Figure \ref{fig:feature_magnitudes}, the amplification is affected as the FAM module modulates feature norms differently for each technique. Given that top-\textit{k} normal and abnormal feature selection is magnitude-based in MGFN, this results in varying model performance, driven by the exposure to differing subsets of events. 

\textbf{UR-DMU} exhibits robustness across most anonymization types, which is attributed to the enforced separability by the dual-memory architecture that stores normal and anomaly instances in different memories. This ensures that each memory unit learns only the relevant (normal or abnormal) patterns. This approach enables preservation of discriminative features despite the added noise. 

\begin{figure}[h!]
\centering
\subfloat[Arson041 video summary]{\includegraphics[width=1\textwidth, height=0.1\textwidth]{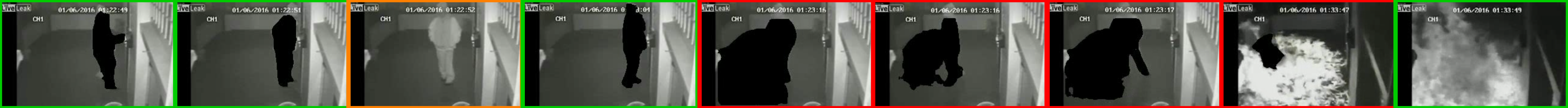}}

\subfloat[Raw ]{\includegraphics[width=0.34\linewidth]{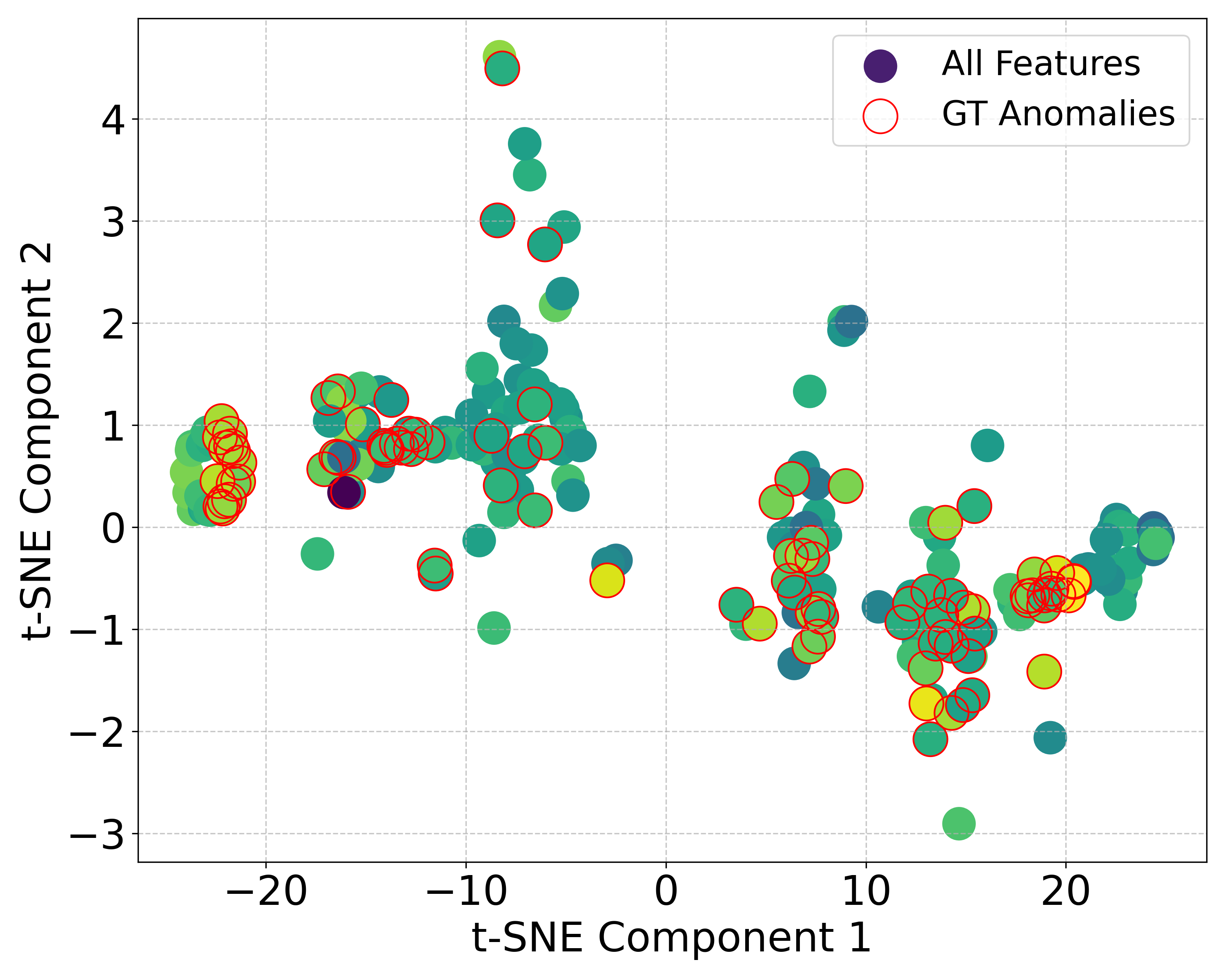}}
\subfloat[HB ]{\includegraphics[width=0.34\textwidth]{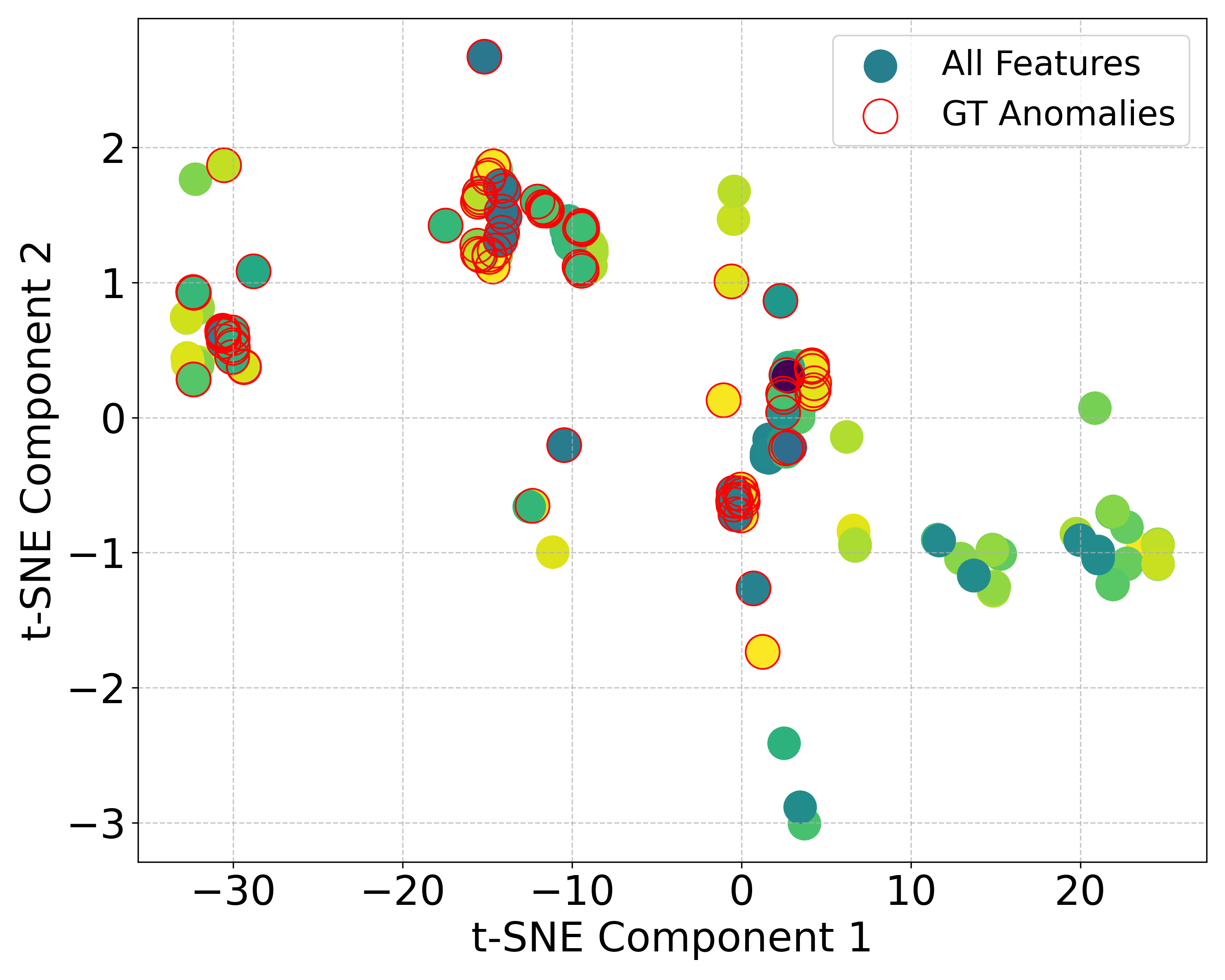}} 
\subfloat[HM ]{\includegraphics[width=0.34\textwidth]{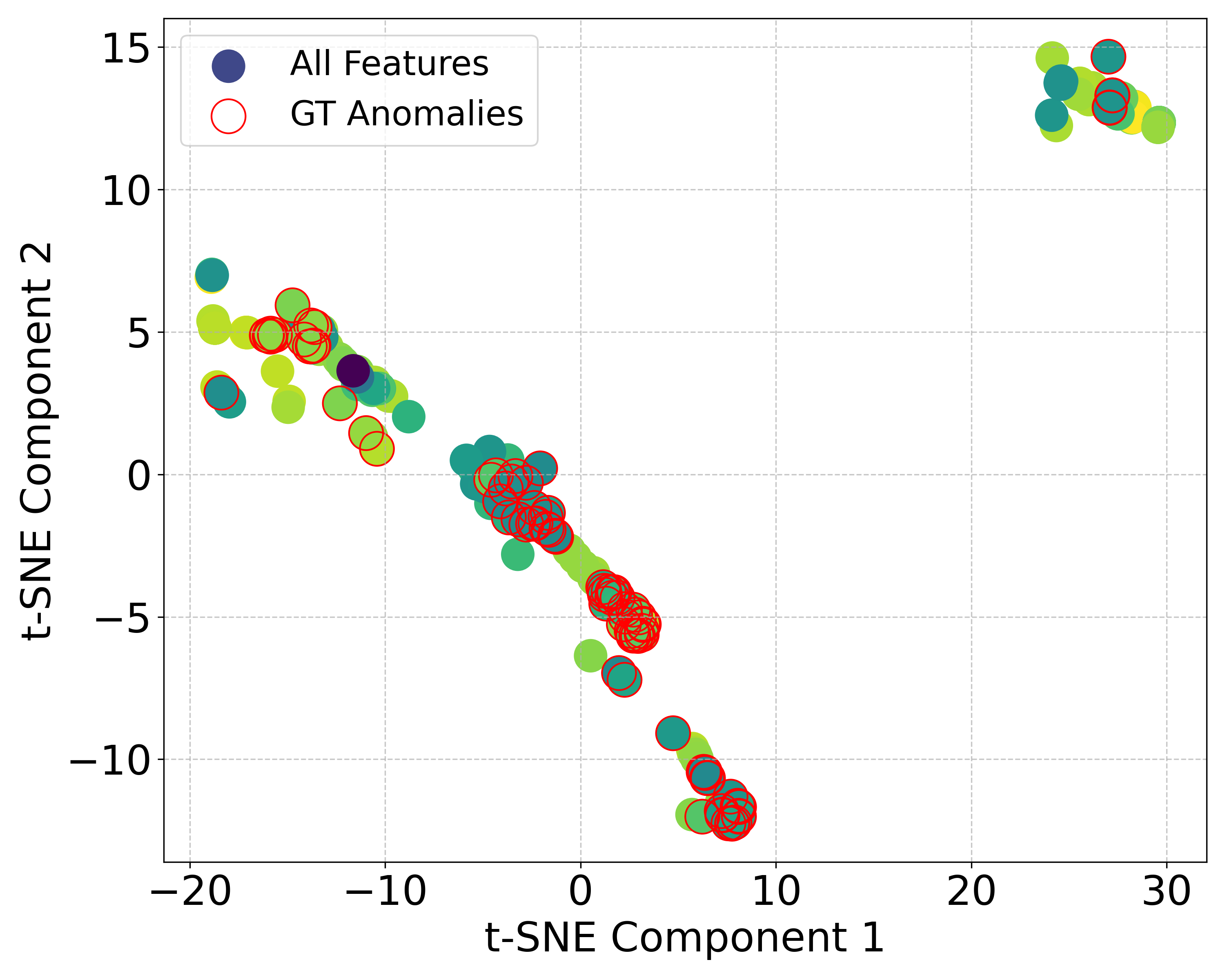}}

\subfloat[HEN ]{\includegraphics[width=0.34\textwidth]{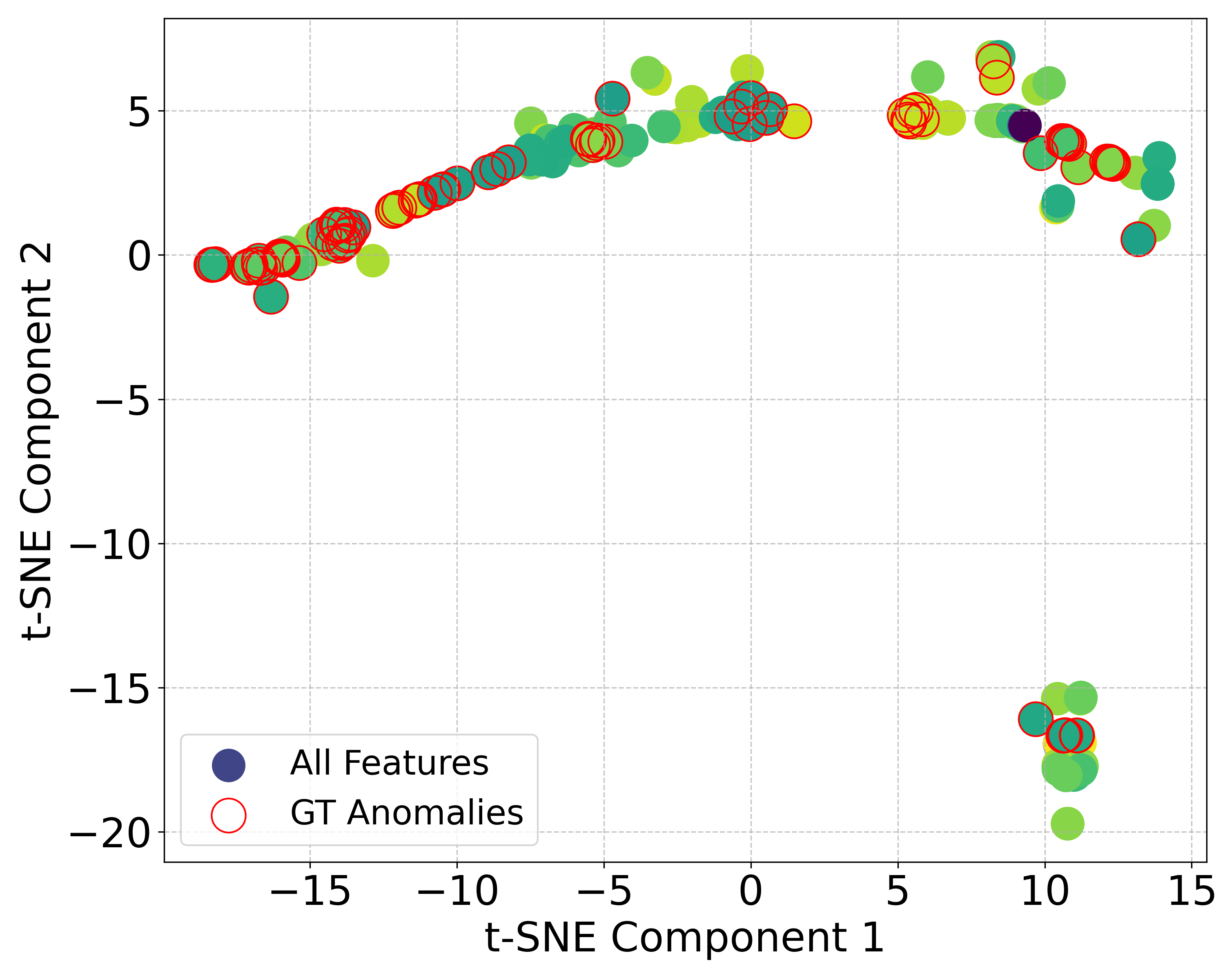}}
\subfloat[H2D ]{\includegraphics[width=0.34\textwidth]{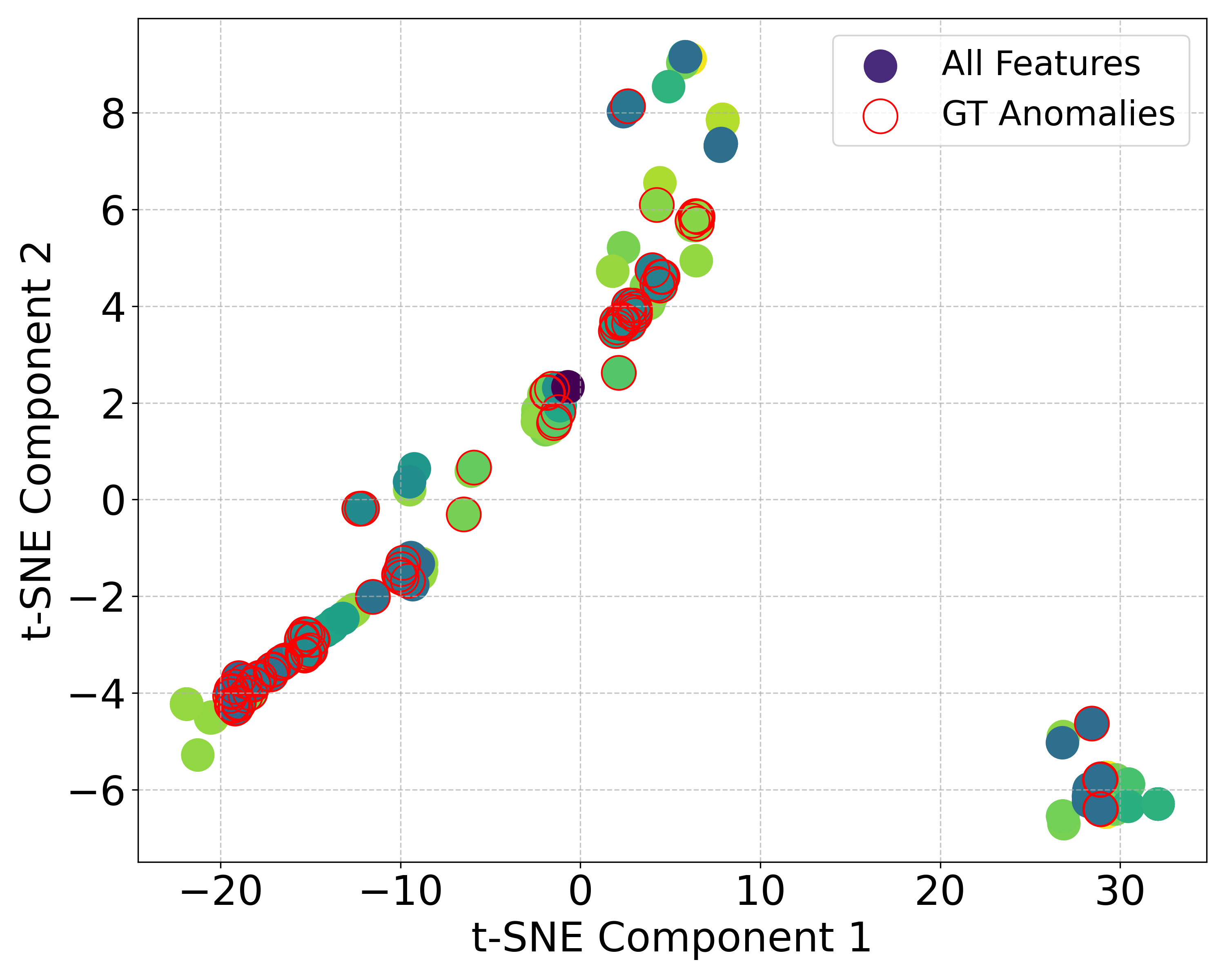}}

\subfloat[Colormap of DFM Score Distribution]{\includegraphics[width=0.8\textwidth]{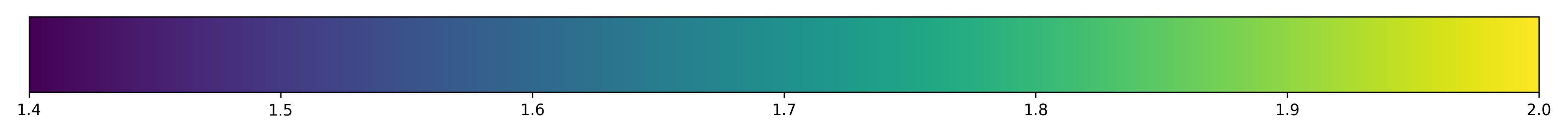}}

\caption{t-SNE visualizations of Arson041 feature snippets, and their divergence from BatchNorm mean (DFM scores) under each anonymization technique (b-f). The DFM scores are computed using the Mahalanobis distance on the enhanced features \cite{zhou2024batchnorm}, and the full video is treated as a batch. Higher DFM scores are represented by lighter colors (g). The video summary in (a) displays frames of the human-masked version, highlighting segmentation flaws in red and orange. These inconsistencies introduce feature variability, increasing the DFM scores for many normal snippets (evident in the lighter points without red circles), and contributing to a higher false alarm rate. }
\label{fig:dfm_examples}
\end{figure}

\textbf{BN-WVAD} demonstrates substantial invariance to human masking, with AUC scores surpassing the performance on non-anonymized data. However, this comes at the cost of a significantly higher false alarm rate (99.65\%). This surprising result may be attributed to the DFM anomaly criterion, which detects deviations from normal by measuring divergence from the BatchNorm-computed mean. Due to the noise introduced by human segmentation, such as inaccurate or missing segments, and the additional distortions from human region replacements, the deviation from BatchNorm statistics becomes an ill-posed criterion. This results in a significant performance fluctuations and a high false alarm rate. The effect is particularly evident in videos with poor anonymization quality, as illustrated in Figure \ref{fig:dfm_examples}. 

\textbf{PEL4VAD} similarly to UR-DMU, exhibits robustness across anonymization types, which is attributed to the explicit modeling of anomalous events context through the prior semantic prompts. These experimental results with PEL4VAD and UR-DMU reveal that without the explicit modeling of nuances that separate the normal and anomaly events during training, the anomaly detection model might end up sensitive to certain patterns or artifacts produced by the anonymization. This sensitivity may arbitrarily boost or degrade the performance, not as a reflection of model effectiveness, but as a reaction to noise-induced variability.

\begin{figure}[h!]
\centering
\subfloat[MGFN \cite{chen2023mgfn}]{\includegraphics[width=0.5\linewidth]{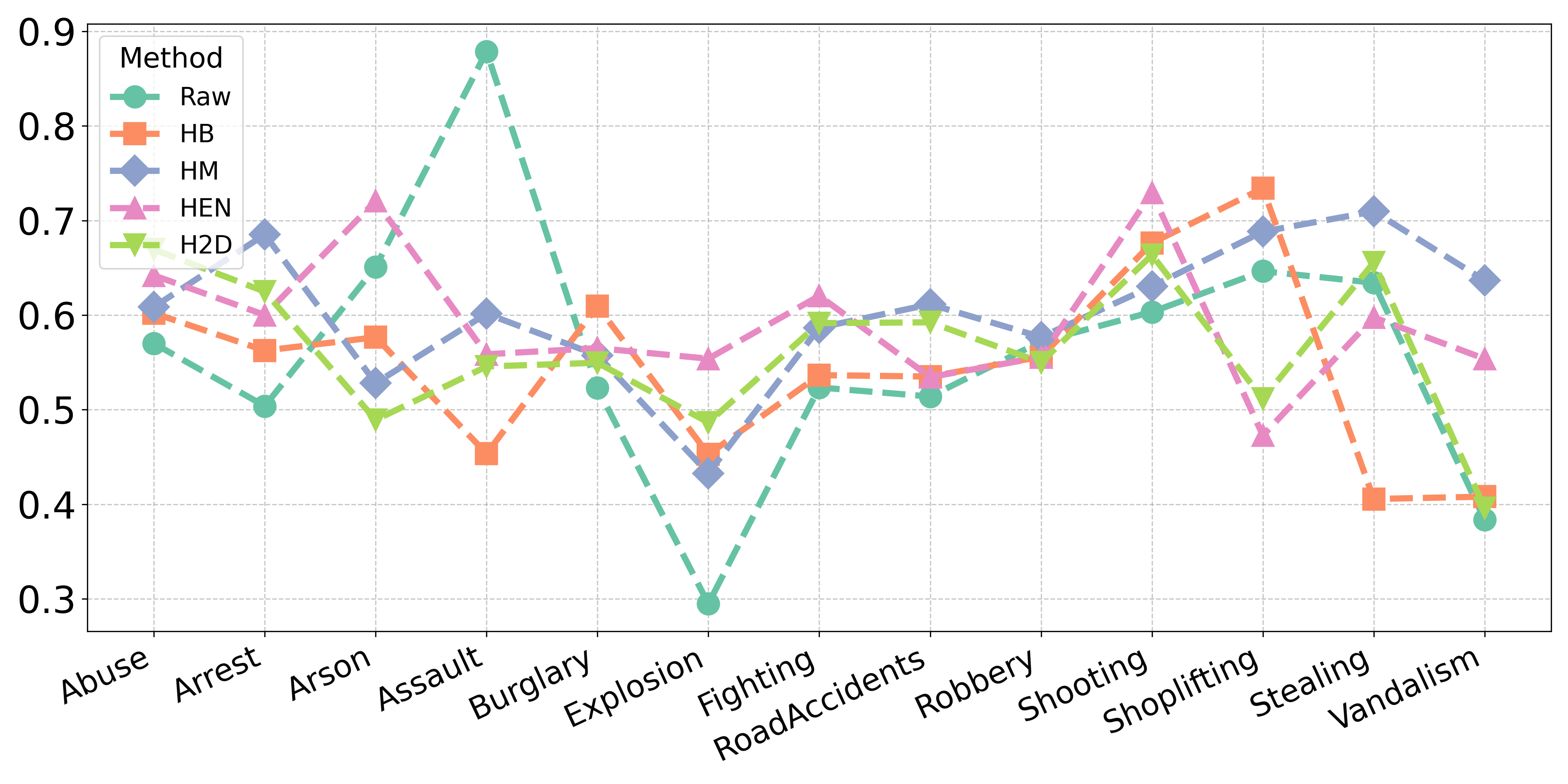}}
\subfloat[UR-DMU \cite{zhou2023dual_urdmu}]{\includegraphics[width=0.5\textwidth]{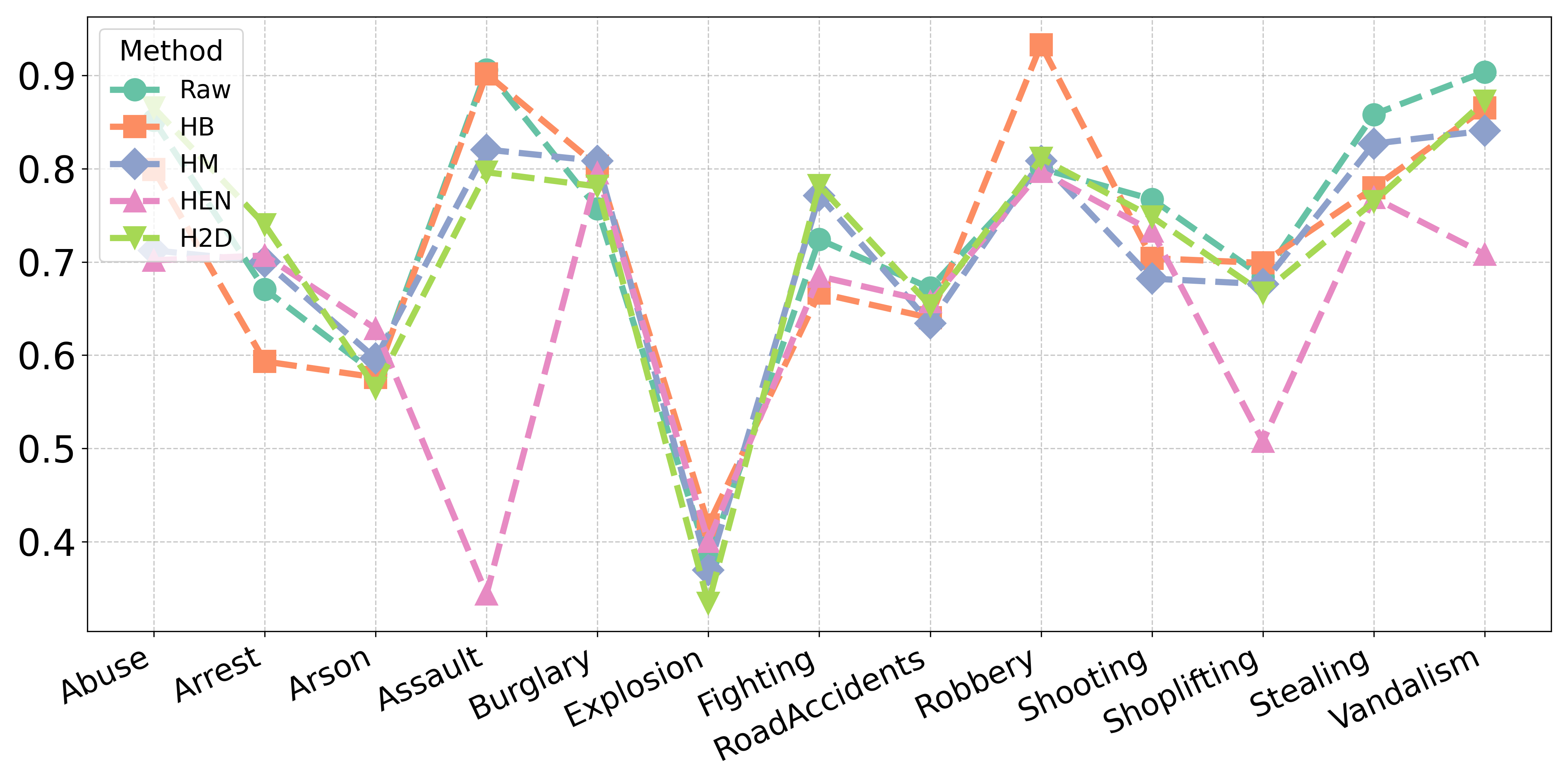}} \hfill
\subfloat[BN-WVAD \cite{zhou2024batchnorm}]{\includegraphics[width=0.5\textwidth]{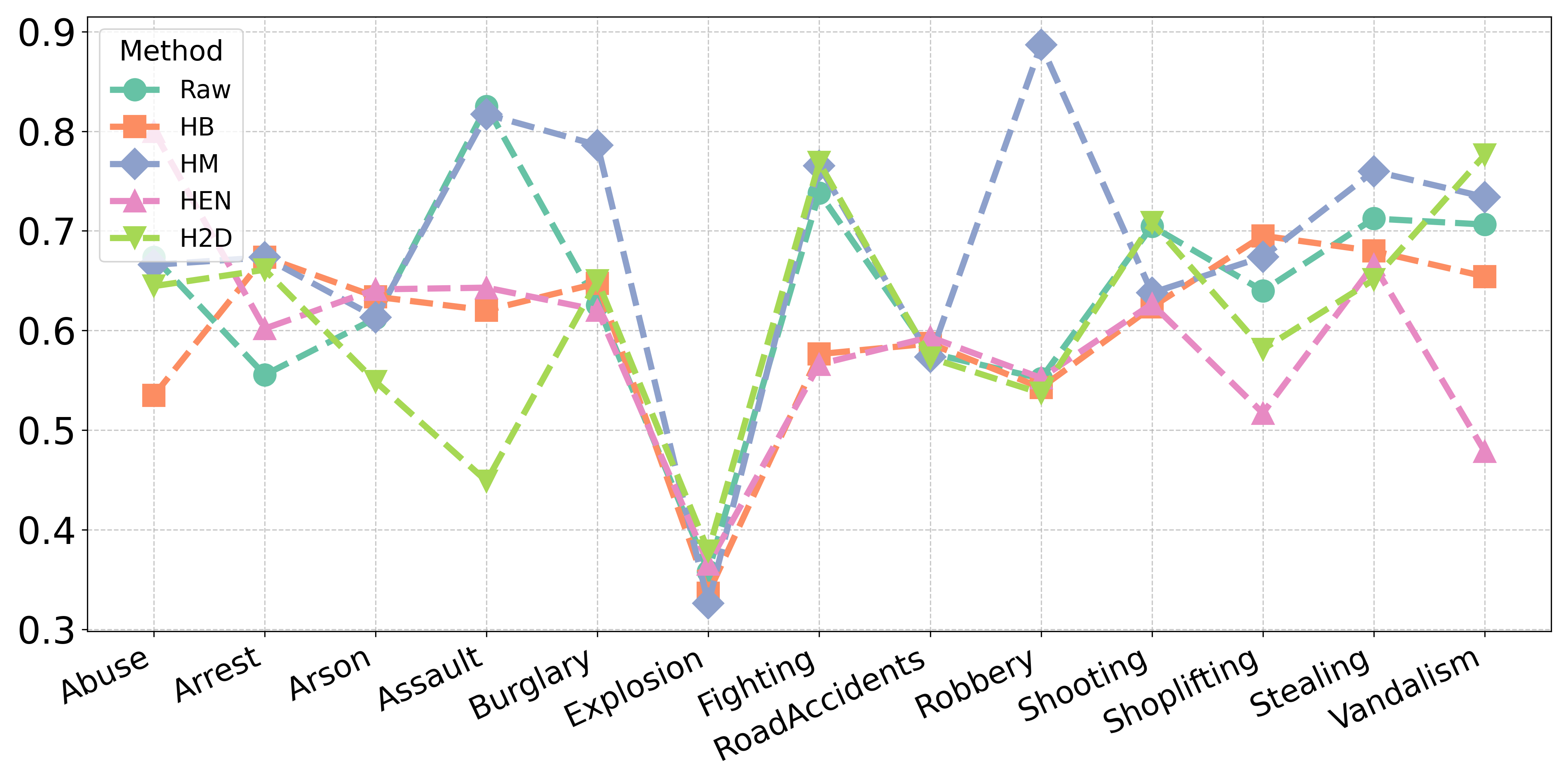}}
\subfloat[PEL4VAD \cite{pu2024learning_pel4vad}]{\includegraphics[width=0.5\textwidth]{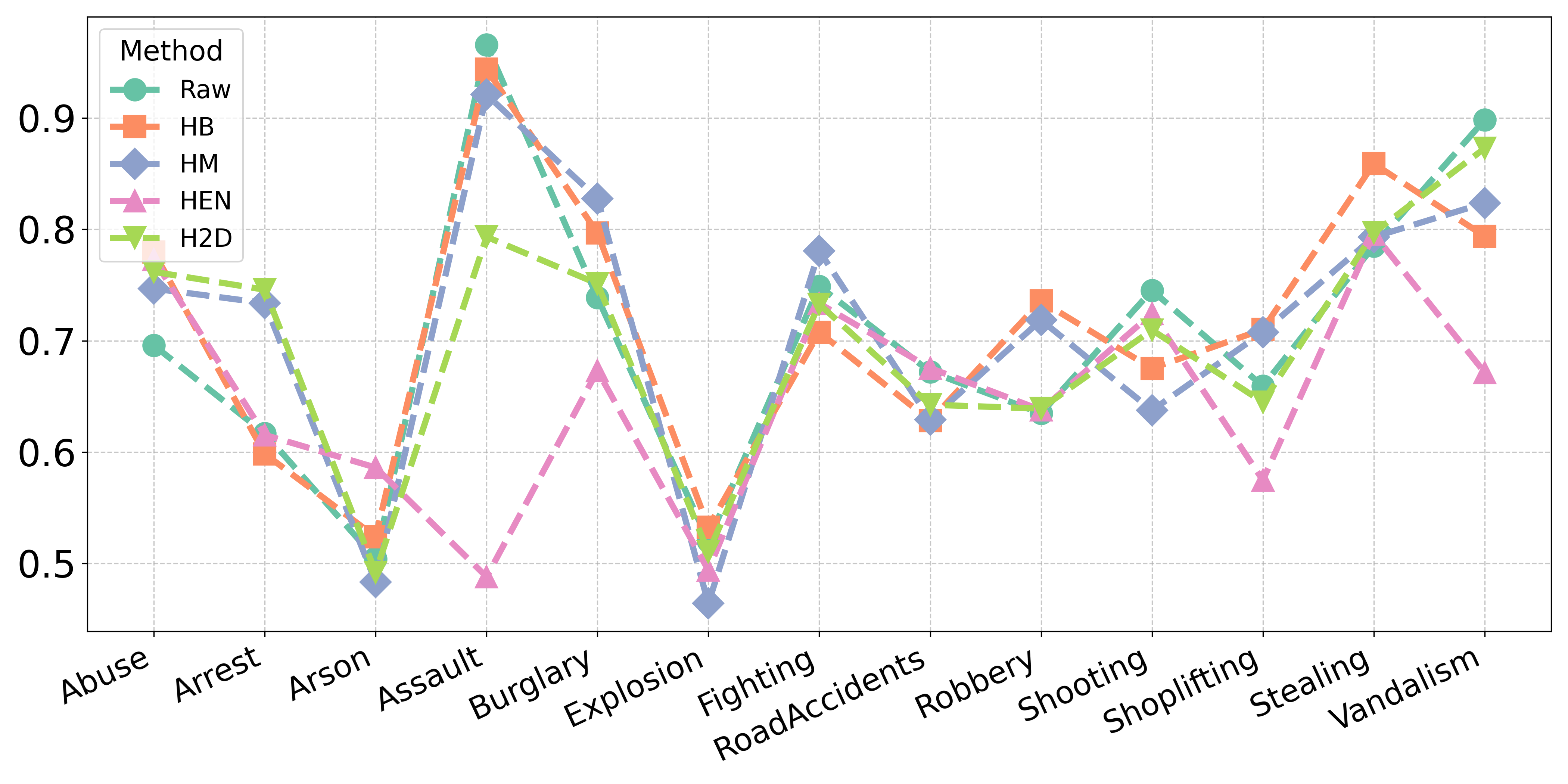}}

\caption{Class-wise AUC performance comparison for the investigated VAD methods. 
}
\label{fig:classwise_scores}
\end{figure}

\subsection{VAD Comparative Analysis}
We further analyse the class-wise AUC for the investigated VAD methods in Figure \ref{fig:classwise_scores}. The UR-DMU exhibits a consistency in the class-wise AUC among the different anonymization techniques. The class-wise AUC scores follow the raw-trained model, except for encryption, which impairs the performance significantly for the ``Assault'', ``Shoplifting'', and ``Vandalism'' classes. This is attributed to the aggressive transformations of encryption, which distort the spatial and temporal information, introducing non-smooth transitions in videos. The lack of temporal smoothness affects the snippet features by disturbing the subtle visual cues essential for accurate identification of these classes. The same observation is seen with the PEL4VAD model on the same classes under the encryption technique, which validates the previous explanation. On the contrary, the MGFN model improves the performance with most anonymization techniques on the classes, such as ``Arrest'', ``Explosion'', ``Fighting'', ``Shoplifting'', ``Stealing'', and ``Vandalism''. This unintentional positive effect of anonymization arises from the dependence on the magnitude amplification technique, as explained earlier, and reflects the model over-sensitivity to noise rather than true effectiveness on context-based anomaly separation. Similarly, the BN-WVAD model shows improvement for some classes particularly under the masking effect. However, these gains are offset by a significantly high false alarm rate. This finding validates the earlier explanation, in which the abnormality criterion of the BN-WVAD model relies on the batch statistics, which are disturbed by the anonymization, leading to a significantly high false alarm rate.

\begin{figure}
    \centering
    \includegraphics[width=\textwidth]{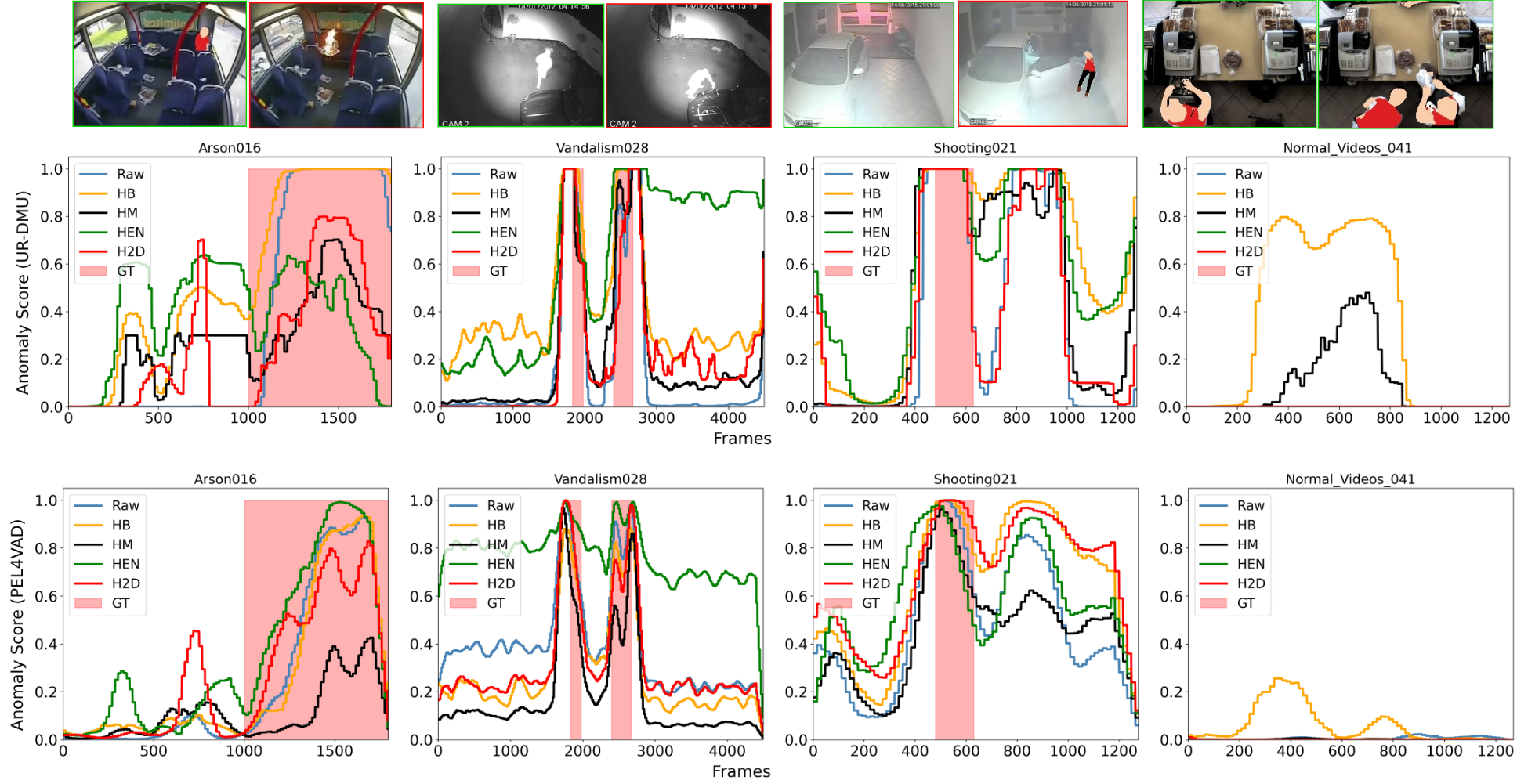}
    \caption{Qualitative results of UR-DMU (top) and PEL4VAD (bottom) anomaly detection under anonymization. }
    \label{fig:anomaly_test_cases}
\end{figure}

\subsection{Qualitative Results}
Figure \ref{fig:anomaly_test_cases} presents qualitative results of the UR-DMU and PEL4VAD models performance across different anonymized versions of AUCF-Crime. Notably, there are significant variations in frame-level anomaly scores across anonymization types, which are attributed to the anonymization artifacts and inaccurate detections between the consecutive frames that act as noise. The variations are more significant for blurring and encryption (HB and HEN), resulting in an increased false alarm rate in normal events. However, in most classes, the anomaly scores closely follow the scores on raw data, which indicates that these anonymization-induced fluctuations can be filtered out by selecting the appropriate threshold. These results highlight the resilience and invariance of the UR-DMU and PEL4VAD models to noise and distortions caused by anonymization, which emphasizes the importance of a VAD method selection for satisfactory performance on anonymized data.

\begin{table}[hbt!]
    \centering
    \caption{Comparison between the best results obtained under conventional anonymization (with $<$1\% performance drop) and privacy-by-design methods (SPAct and TeD-SPAD) when applied to anomaly detection. *indicates that the scores are reported from original papers.  }
    \renewcommand{\arraystretch}{1}
    \setlength{\tabcolsep}{5pt}  % Adjust column spacing
    \begin{tabular}{l|cc}
        \hline
        \textbf{Method} & AUC(\%)($\uparrow$) & Relative drop (\%)($\downarrow$) \\
        \hline
         UR-DMU \cite{zhou2023dual_urdmu} (Raw-Ours) & \textbf{85.85} & -  \\
         PEL4VAD \cite{pu2024learning_pel4vad} (Raw-Ours) & 85.16 & -  \\
         \hline
         SPAct \cite{fioresi2023ted}* & 73.93 & \textcolor{red}{$\downarrow$ 4.83 \%}   \\
         TeD-SPAD \cite{fioresi2023ted}* & \textbf{74.81} & \textbf{\textcolor{red}{$\downarrow$ 3.69 \%}}      \\
        \hline
        UR-DMU-HB  & \textbf{85.63} & \textbf{\textcolor{red}{$\downarrow$ 0.25\%}}  \\
        UR-DMU-HM & 85.08 & \textcolor{red}{$\downarrow$ 0.89 \%}  \\
        PEL4VAD-HB  & 84.72 & \textcolor{red}{$\downarrow$ 0.51 \%}  \\
        PEL4VAD-HM & 84.53 & \textcolor{red}{$\downarrow$ 0.74 \%}  \\
        \hline
    \end{tabular}
    \label{tab:compare_pbd_solutions}
\end{table}

\subsection{Privacy-by-Design vs. Conventional Anonymization} 
We present the comparative analysis between the conventional anonymization techniques in this study and recent privacy-by-design solutions, SPAct and TeD-SPAD \cite{dave2022spact,fioresi2023ted}, in Table \ref{tab:compare_pbd_solutions}. These solutions employ a weakly supervised anomaly detection approach, based on the MGFN \cite{chen2023mgfn}. It is shown that, despite being more protective, privacy-by-design methods exhibit a higher performance drop in comparison to conventional anonymization techniques.

It is important to highlight a key methodological difference compared to evaluations in these previous works. In the original privacy-by-design studies, comparisons with blurring and blackening were performed based on bounding-box based anonymization. This presents the extreme form of human obfuscation that removes significant shape information and contextual details. In contrast, our evaluation employs human-segment anonymization, preserving the crucial shape and contextual information. Consequently, the results indicate that human-segment anonymization can achieve higher AUC scores and lower performance drops than those reported by privacy-by-design methods.

Conventional anonymization techniques provide a lower privacy protection degree compared to privacy-by-design solutions. However, they are shown to maintain satisfactory anomaly detection performance, and are flexible to other forms of image processing algorithms \cite{eval-framework}. On the contrary, even though privacy-by-design solutions offer optimal protection, they compromise utility to a higher degree than the privacy gain. In addition, they exhibit poor applicability of their data to generic image processing tasks, due to the massive information loss. These observations imply an essential yet often overlooked trade-off in evaluating the privacy-by-design approaches, which is the balance between achieving robust privacy protection and retaining broad applicability across diverse utility tasks.

\section{Conclusion}
In this paper, we have analyzed the robustness of multiple anomaly detection methods under four human anonymization techniques that selectively mask, distort, or replace sensitive body regions. Our experiments demonstrate that conventional anonymization does not inherently prevent anomaly detection. Some VAD methods, such as MGFN with encryption or BN-WVAD with masking, can inadvertently show performance gain in response to the anonymization artifacts, whereas UR-DMU and PEL4VAD show robustness across all anonymization types. The evaluations indicate that the performance of anomaly detection models, that employ I3D features under anonymization, highly depends on the model design and the learning strategy, suggesting that neither anonymity nor accuracy need to be entirely sacrificed. Furthermore, when comparing the conventional anonymization techniques with privacy-by-design solutions, the analysis reveals an often-neglected trade-off between the application flexibility and privacy protection. Conventional anonymization can preserve scene context for multiple downstream tasks, whereas privacy-by-design methods typically optimize for a single target utility with compromised performance. We expect these findings to encourage further comparative investigations into how VAD methods handle anonymization artifacts and how much they rely on sensitive human details. We also aim to inspire new methods that ensure robust privacy while maintaining generalizable feature representations.

\bibliography{references}

% \appendix
\end{document}